\documentclass{article}

\usepackage{microtype}
\usepackage{graphicx}
\usepackage{subfigure}
\usepackage{booktabs} 
\usepackage{amsmath,amsthm,epsfig,amsfonts,amssymb}
\usepackage{graphics}
\usepackage{amstext}
\usepackage{color}
\usepackage{float}
\usepackage{natbib}
\usepackage{appendix}
\usepackage{caption}

\usepackage{multicol}
\usepackage[export]{adjustbox}

\usepackage{hyperref}

\usepackage[accepted]{icml2021}

\icmltitlerunning{Generation and Simulation of Synthetic Datasets with Copulas}

\theoremstyle{definition}

\begin{document}
\twocolumn[
\icmltitle{Generation and Simulation of Synthetic Datasets with Copulas}

\begin{icmlauthorlist}
\icmlauthor{R\'egis Houssou}{heig}
\icmlauthor{Mihai-Cezar Augustin}{heig}
\icmlauthor{Efstratios Rappos}{heig}
\icmlauthor{Vivien Bonvin}{ng}
\icmlauthor{Stephan Robert-Nicoud}{heig}
\end{icmlauthorlist}

\icmlaffiliation{heig}{Haute Ecole de la Suisse Occidentale (HES-SO), Switzerland}
\icmlaffiliation{ng}{NetGuardians SA}
\icmlcorrespondingauthor{}{stephan.robert@hes-so.ch}
\icmlkeywords{Machine Learning, ICML}

\vskip 0.3in
]

\printAffiliationsAndNotice{} 

\begin{abstract}
\noindent 

This article proposes a new method to generate synthetic data sets based on copula models. Our goal is to produce surrogate data resembling real data in terms of marginal and joint distributions. We present a complete and reliable algorithm for generating a synthetic data set comprising numeric or categorical variables. Applying our methodology to two datasets shows better performance compared to other methods such as SMOTE and autoencoders.
\end{abstract}

\section{Introduction}
Governments, organizations, businesses, academia, members of the public and other decision-making bodies all would like easier and unrestricted access to good quality data to make informed and accurate decisions. However, accessing a good, real-world dataset can be difficult, expensive, time-consuming, or simply impossible, and real data is almost always vulnerable to data privacy breach incidents. These data-limited situations can limit the efforts of the data science community to build robust models for organizations to make informed policy decisions and address major challenges, \cite{loan}.

In this work, we propose to use synthetic data generation methods to produce surrogate data closely resembling real data in data limited situations, thus minimizing the need to access real data. Synthetic data is a powerful tool when the data required is limited or when it is desired to be shared securely with the parties concerned. The idea of synthetic data, that is, data artificially fabricated rather than obtained by direct measurement, was introduced by \cite{rubin}, who used multiple imputation to generate a synthetic version of the Decennial Census. As a result, he was able to release samples without disclosing microdata.

\cite{rubin}  first attempted to generate synthetic data in 1993 using multiple imputation, for which parametric modeling was used.  \cite{reiter} points out, however, that it is generally difficult and tedious to produce precise models based on standard parametric approaches. Therefore, nonparametric machine learning approaches have been used, such as classification and regression trees, support vector machines, bagging, and random forests, see  \cite{loan}. With recent developments in deep learning, deep neural networks have started to become an attractive option for generating synthetic data, given their ability to use big data to create powerful models. The main deep learning algorithms, widely used for data synthesis, especially in the fields of image and video, are antagonistic generator networks (GAN), \cite{goodf} And autoencoders \cite{ballard}.

In this paper, we propose to use a robust statistical model such as copulas \cite{Nel} to generate synthetic data sets that closely resemble real data. Copulas are a powerful tool for isolating the structure of dependencies in a multivariate distribution. It is a probability model that represents a multivariate uniform distribution, which examines the association or dependence between many variables. Using copulas and the inverse transformation method, we can produce a synthetic data set that mimics the real data in terms of association or dependence and marginal distribution. In addition, our copula-based method can be used for imputation (replacing missing data with substituted values) and augmentation of real data with synthetic data, to build robust statistical, machine-learning and deep-learning models more rapidly and at lower cost. Another challenge in the present work is to generate a synthetic dataset from an actual dataset that includes categorical features. In this context, we propose an algorithm that converts categorical features to numeric ones, then generates the synthetic copula-based dataset and finally converts the generated features to categorical ones. Thus, our proposed method is a complete and robust algorithm to generates a synthetic dataset including numeric or categorical features, which makes possible to share data more easily and more quickly between government, universities and the private sector.

The rest of the paper is organized as follows. Section II gives an overview of copulas and the algorithm to simulate them. We also present the rank correlation and the relationship with copulas. In Section III, our approach of generating synthetic datasets with numerical and categorical features is presented. Section IV presents an application to datasets and the comparison with other approaches such as autoencoders-based and the Synthetic Minority Oversampling (SMOTE) approach. The conclusion is drawn in Section V.
\section{Overview of copulas}
Copula \cite{Nel} is defined as a probability distribution over a high-dimensional unit cube $\left[0,1 \right]^{n}$ whose univariate marginal distributions are uniform on $\left[0,1 \right]$. Copula can model the dependencies of large dimensional random variables and is very efficient in risk management or other tasks that require dependency modelling, see \cite{mcN}, \cite{kole}, etc...  Formally, given a set of uniformly distributed random variables, $U_{1},U_{2},...,U_{n}$, a copula is a joint cumulative distribution defined as 
\begin{equation}
C(u_{1},u_{2},...,u_{n})=P(U_{1}\leq u_{1},...,U_{n}\leq u_{n})
\end{equation}
with the following properties 
\begin{enumerate}
\item $C(u_{1},u_{2},...,u_{n})$ in non-decreasing in each component, $u_{i}$
\item The ith marginal distribution is obtained by setting $u_{j}=1$ for $j \neq i $ and since it is uniformly distributed  $$ C(1,...,1,u_{i},1,...1)=u_{i}$$
\item For $a_{i} \leqslant b_{i}$, $P(U_{1} \in \left[a_{1},b_{1} \right],...,U_{n} \in \left[a_{n},b_{n} \right])$ must be non-negative.
\end{enumerate}

The reverse is also true in that any function that satisfies properties $1$ to $3$ is a copula. 

What makes a copula model above so useful is \textbf{Sklar's} famous theorem \cite{sklar}.  It states that for any joint cumulative distribution function (CDF) with a set of continuous random variables $\lbrace X_{i}\rbrace_{1}^{n}$ and marginal CDF $F_{i}(x_{i})=P(X_{i} \leq x_{i})$, there exists a unique copula function such that the joint CDF is 
\begin{equation}
F(x_{1},...,x_{n})=C(F_{1}(x_{1}),...,F_{n}(x_{n}))
\label{lcop}
\end{equation}
By probability integral transform, each marginal CDF is a uniform random variable on $\left[0,1 \right]$. Therefore, the above copula is valid. In the opposite direction, consider a copula C, and marginal CDF, $F_{i}(x_{i})$. Then, $F$ as defined in (\ref{lcop}) is a multivariate CDF with marginals $F_{i}(x_{i})$. 
And, for a given joint distribution, we can also find the corresponding copula which is the CDF function of the given marginals.

A useful representation that we can get by Sklar's theorem for a continuous copula is,
\begin{equation}
C(u_{1},...,u_{n})=F(F_{1}^{-1}(u_{1}),...,F_{n}^{-1}(u_{n}))
\label{lcop1}
\end{equation}
One of the most important result in the theory of copula is the theorem of \textbf{Invariance Under Monotonic Transformations}.  It basically states that if we apply a monotonic transformation to each component of $X=(X_{1},...,X_{n})$ then the copula of the resulting multivariate distribution remains the same.

Consider a copula $C(u)=C(u_{1},...,u_{n})$; by the theorem of \textbf{Fr\'echet-Hoeffding bounds}, we have the following result 
\begin{equation}
\text{max}\left\lbrace  1-n-\sum_{i=1}^{n}u_{i},0  \right\rbrace  \leq C(u)\leq \text{min} \left\lbrace u_{1},...,u_{n}\right\rbrace
\end{equation}
The upper Fr\'echet-Hoeffding bound is tight for all $n$ while the lower Fr\'echet-Hoeffding Bound is tight only when $n=2$.  These bounds correspond to cases of extreme of dependency, i.e. comonotonicity and countermonotonicity. The comonotonic copula is given by 
\begin{equation}
M(u)=\text{min} \left\lbrace u_{1},...,u_{n}\right\rbrace
\end{equation}
which is the Fr\'echet-Hoeffding upper bound. This corresponds to the case of extreme positive dependence.
The countermonotonic copula is the 2-dimensional copula that is the Fr\'echet-Hoeffding lower Bound. It is given by 
\begin{equation}
W(u)=\text{max}\left\lbrace  u_{1}+u_{2}-1,0  \right\rbrace
\end{equation}
and corresponds to the case of perfect negative dependence.

The independence copula satisfies 
\begin{equation}
\Pi(u)=\prod_{i=1}^{n} u_{i}
\end{equation}
and this corresponds to the case of independence random variables.

There are other families of copulas such as elliptical copulas (Gaussian copula, t copula), Archimedean copulas (Gumbel copula, Clayton copula), Extreme Value copulas, etc., see \cite{mcN}, \cite{jo} and \cite{nelsen}. In this study, we will focus on the elliptical copulas such as Gaussian and t copulas that are used in this paper to generate synthetic dataset. 

\subsection{Elliptical copulas: Gaussian and t copulas}
\label{pop}
\subsubsection{Definition and simulation of Gaussian copulas}
Recall that when the marginal CDFs are continuous, we have from (\ref{lcop1}) that
$$C(u_{1},...,u_{n})=F(F_{1}^{-1}(u_{1}),...,F_{n}^{-1}(u_{n}))$$
Suppose that $X \sim \mathcal{\text{MN}}(0,P)$ with $P$ is the correlation matrix of $X$. then the corresponding Gaussian copula is defined as 
\begin{equation}
C_{P}^{\text{Gau}}(u_{1},...,u_{n})=\phi_{P}(\phi^{-1}(u_{1}),...,\phi^{-1}(u_{n}))
\label{lcop2}
\end{equation}
where $\phi$ is the standard normal cumulative distribution and $\phi_{P}$ the joint CDF of $X$. For imperfectly correlated variables, the Gaussian copula implies tail independence.

\medskip

\textbf{Simulation of Gaussian Copulas}

For an arbitrary covariance matrix $\Sigma$, let $P$ be its corresponding correlation matrix
\begin{enumerate}
\item Compute the Cholesky decomposition, $A$ of $P$ so that $P=AA^{T}$ with $A^{T}$ the transpose of $A$.
\item Generate $Z \sim \mathcal{\text{MN}}(0,I)$ with $I$ being the identity matrix.
\item Set $W=AZ$. In fact, $W \sim \mathcal{\text{MN}}(0,P)$
\item Return $U=(\phi(W_{1}),...,\phi(W_{n}))$. The distribution of $U$ is the Gaussian copula $C_{P}^{\text{Gau}}$. 
\end{enumerate}

\subsubsection{Definition and simulation of t copulas}
The general multivariate $t$ distribution $X$, with $\nu$ the degree of freedom (d.o.f), $\mu$ the location vector and $\Sigma$ the scale matrix can be defined by the stochastic representation 

\begin{equation}
X=\mu+\sqrt{V}BZ
\end{equation}

where
\begin{equation*}
V=\frac{\nu}{\chi_{\nu}^{2}}
\end{equation*}

\begin{figure}
\begin{multicols}{2}
\includegraphics[width=\linewidth]{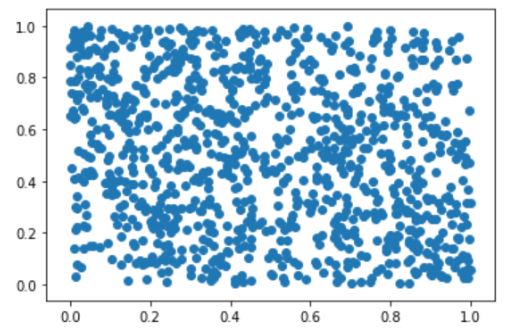}\par 
\includegraphics[width=0.97\linewidth]{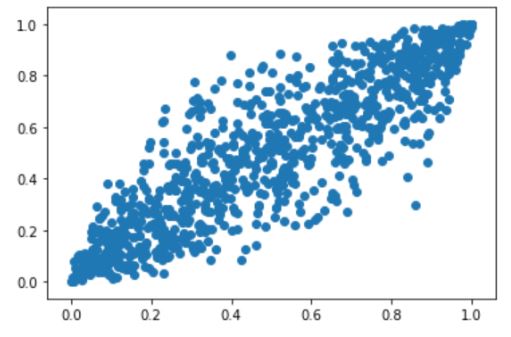}\par
\includegraphics[width=0.93\linewidth]{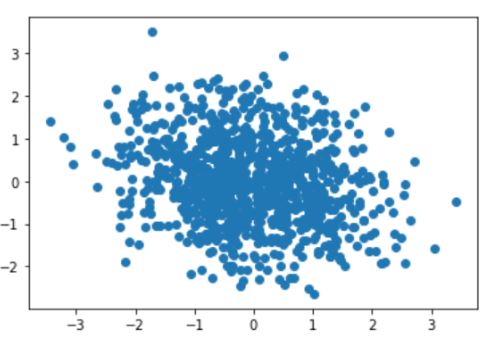}\par 
\includegraphics[width=0.93\linewidth]{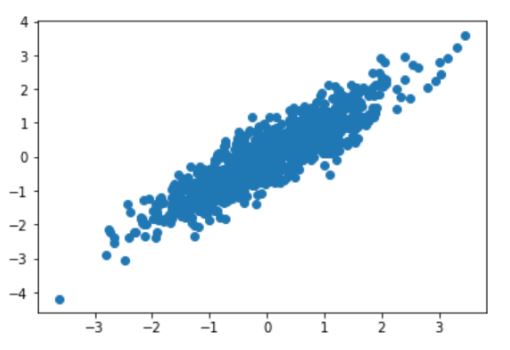}\par
\end{multicols}
\caption{Simulation of Gaussian copula and bivariate Gaussian samples. (top left) Bivariate Gaussian Copula, $\rho=-0.2$. (top right) Bivariate Gaussian random samples, $\rho=-0.2$. (bottom left) Bivariate Gaussian Copula, $\rho=0.9$. (bottom right) Bivariate Gaussian random samples, $\rho=0.9$.}
\label{fig:fig1}
\end{figure}

$\chi_{\nu}^{2}$ follows a chi-squared distribution with $\nu>0$ and is independent of $Z \sim \text{MN}(0,I)$; $B$ is such that $BB^{T}=\Sigma$. We denote $X \sim t_{\nu,\mu,\Sigma}$; for more details, see \cite{hofert}, \cite{shaw}, etc.

Suppose that $X=(X_{1},...,X_{n})\sim t_{\nu,0,P}$ where $\nu>2$ and $P$ its correlation matrix. The corresponding $t$-copula is defined as

\begin{equation}
C_{\nu,P}^{t}=t_{\nu,0,P}(t_{\nu}^{-1}(u_{1}),...,t_{\nu}^{-1}(u_{n}))
\end{equation}

where $t_{\nu}$ is the univariate CDF of $t$-distribution with mean zero and dispersion parameter equal $1$ and d.o.f, $\nu>2$.

A higher value for $\nu$ decreases the probability of tail events. When the $t$-copula converges to the Gaussian copula for $\nu \rightarrow \infty$, the $t$ copula assigns more probability to tail events than the Gaussian copula. In addition, the $t$-copula exhibits tail dependence even if the correlation coefficients are equal to zero.

\medskip

\textbf{Simulation of t Copulas}

\begin{figure}
\begin{multicols}{2}
\includegraphics[width=\linewidth]{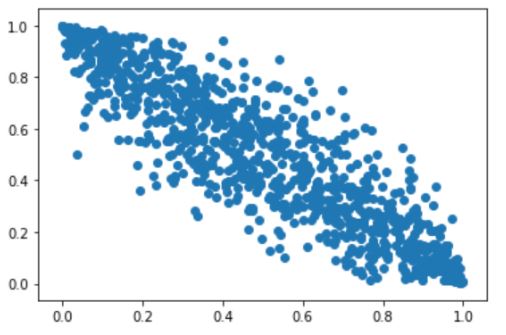}\par 
\includegraphics[width=\linewidth]{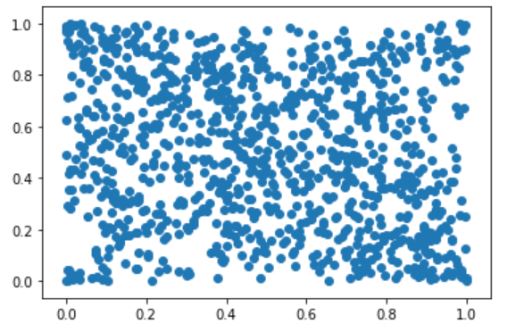}\par 
\includegraphics[width=\linewidth]{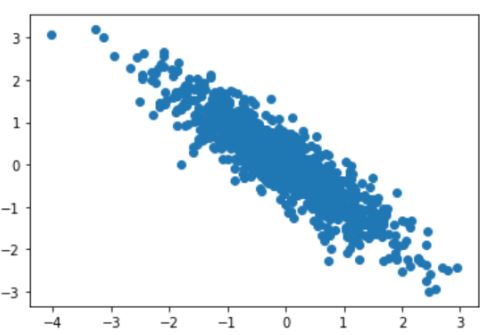}\par
\includegraphics[width=\linewidth]{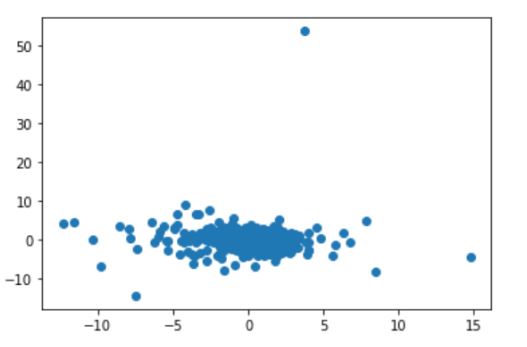}\par
\end{multicols}
\caption{Simulation of Gaussian copula, bivariate Gaussian samples, $t$-copula and bivariate $t$ random samples  (top left) Bivariate Gaussian Copula, $\rho=-0.9$. (top right) Bivariate Gaussian random samples, $\rho=-0.9$. (bottom left) Bivariate $t$-Copula, $\rho=-0.2$, $\nu=3$. (bottom right) Bivariate $t$ random samples, $\rho=-0.2$, $\nu=3$.}
\label{fig:fig2}
\end{figure}

\begin{figure}
\begin{multicols}{2}
\includegraphics[width=\linewidth]{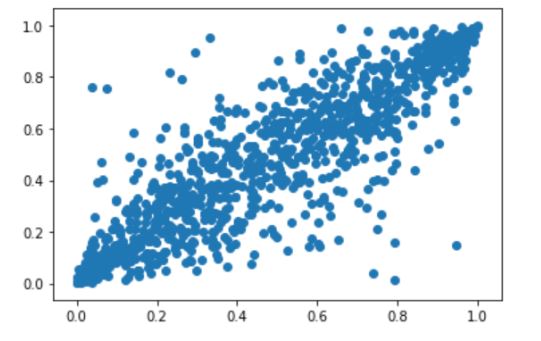}\par 
\includegraphics[width=\linewidth]{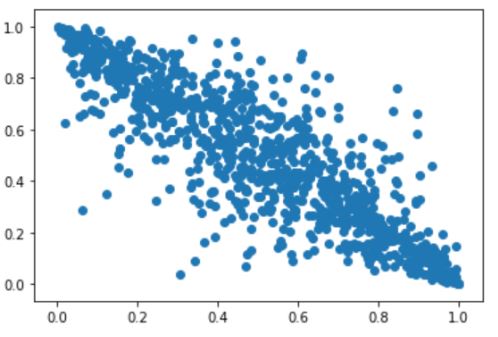}\par 
\includegraphics[width=\linewidth]{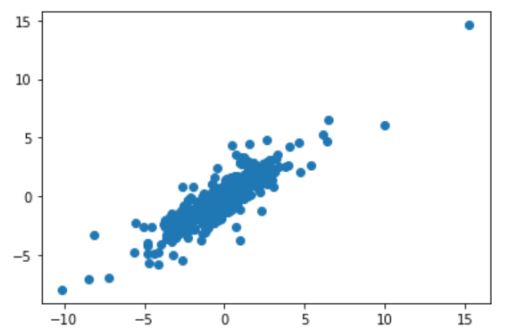}\par
\includegraphics[width=\linewidth]{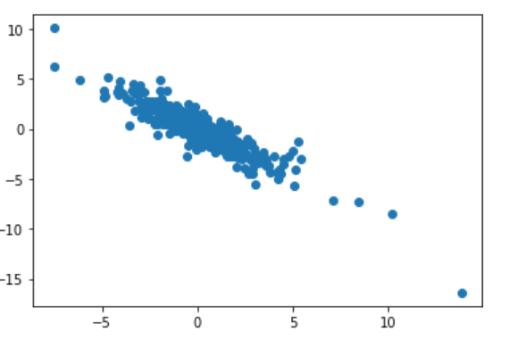}\par
\end{multicols}
\caption{Simulation of $t$-copula and the bivariate $t$ random samples. (top left) Bivariate $t$-Copula, $\rho=0.9$, $\nu=3$. (top right) Bivariate $t$ random samples, $\rho=0.9$, $\nu=3$. (bottom left) Bivariate $t$-Copula, $\rho=-0.9$, $\nu=3$. (bottom right) Bivariate $t$ random samples, $\rho=-0.9$, $\nu=3$}
\label{fig:fig3}
\end{figure}

Let $\Sigma$ be the covariance matrix of the input data $X=(X_{i},...,X_{n})$ and $P=cor(X)$ its corresponding correlation matrix. The steps to generate the t copulas are following:
\begin{enumerate}
\item Generate  $W \sim \mathcal{\text{MN}}(0,P)$
\item For $\nu>2$, generate $\epsilon \sim \chi_{\nu}^{2}$ independent of $W$
\item Return $U=\left( t_{\nu}\left(\frac{W_{1}}{\sqrt{\frac{\epsilon}{\nu}}}\right),...,t_{\nu}\left(\frac{W_{n}}{\sqrt{\frac{\epsilon}{\nu}}}\right)\right) $, where $t_{\nu}$ is the CDF of a standard univariate $t$ distribution with d.o.f, $\nu$. The distribution of $U$ is in fact the $t$-copula $C_{\nu,P}^{t}$
\end{enumerate}
Figures \ref{fig:fig1} to \ref{fig:fig3} show the simulations of the bivariate Gaussian and $t$-copulas with the associated random samples using the above algorithms. We consider different correlation coefficients such as $\rho=-0.2$, $\rho=0.9$ and $\rho=-0.9$.

\subsubsection{Rank Correlation and copulas} 
In this section, we briefly present the Rank correlations which provide the best alternatives to the linear correlation coefficient as a measure of dependence for non-elliptical distributions, for which the linear correlation coefficient is inappropriate and often misleading. In addition, we provide the relation existing between the Rank correlation and the copula and a moment-matching approach for fitting either the Gaussian or t copulas.  There are two important rank correlation measures, namely Spearman's rho and Kendall's tau.  

\medskip
\newpage

\textbf{Spearman's rho}

Let's $X_{1}$ and $X_{2}$ two random variables and $\rho$ the linear (Pearson) correlation coefficient. The Spearman's rho is defined as 
\begin{equation}
\rho_{s}(X_{1},X_{2})=\rho(F_{1}(X_{1}),F_{2}(X_{2}))
\end{equation} 
Spearman's rho is the linear correlation of the probability-transformed random variables. If $X_{1}$ and $X_{2}$ have continuous marginals, it can be shown the following relationship between the Spearman's rho and the unique copula $C$ 
\begin{equation}
\rho_{s}(X_{1},X_{2})=12 \int^{1}_{0} \int^{1}_{0} (C(u_{1},u_{2})-u_{1}u_{2}) du_{1}du_{2}
\label{spear}
\end{equation}
In the case of bivariate Gaussian copula, (\ref{spear}) is reduced to 

\begin{equation}
\rho_{s}(X_{1},X_{2})=\frac{6}{\pi} \text{arcsin} \left( \frac{\rho}{2} \right)  \approx \rho
\end{equation}


\medskip

\textbf{Kendall's tau}

Let's $X_{1}$ and $X_{2}$ two random variables. The Kendall's tau is defined as 
\begin{equation}
\rho_{\tau}\left(X_{1},X_{2}\right)=E\left[\text{sign}\left((X_{1}-X_{1}^{*})(X_{2}-X_{2}^{*})\right)\right]
\label{ken}
\end{equation}
where $(X_{1}^{*}, X_{2}^{*})$ is independent of $(X_{1}, X_{2})$ but has the same joint distribution as $(X_{1}, X_{2})$. Kendall's tau can also be written as 
\begin{equation*}
\begin{aligned}
\rho_{\tau}\left(X_{1},X_{2}\right)&= P\left((X_{1}-X_{1}^{*})(X_{2}-X_{2}^{*})>0\right)- \\
&P\left((X_{1}-X_{1}^{*})(X_{2}-X_{2}^{*})<0\right)
\end{aligned}
\end{equation*}
Kendall's tau is simply the probability of concordance minus the probability of discordance. So if both probabilities are equal then $\rho_{\tau}\left(X_{1},X_{2}\right)=0$. If $X_{1}$ and $X_{2}$ have continuous marginals, it can be shown that
\begin{equation}
\rho_{\tau}\left(X_{1},X_{2}\right)= 4 \int^{1}_{0} \int^{1}_{0} C(u_{1},u_{2})\,\, dC(u_{1},u_{2})-1
\end{equation}
It may also be shown that for a bivariate Gaussian copula
\begin{equation}
\rho_{\tau}\left(X_{1},X_{2}\right)= \frac{2}{\pi} \text{arcsin} \left( \rho\right) 
\label{ken1}
\end{equation}
where $\rho$ is the linear (Pearson) correlation coefficient. (\ref{ken1}) can be very useful to estimate $\rho$ for fat-tailed elliptical distribution such as the $t$-distribution. This has been experienced by \cite{martinH} who showed that (\ref{ken1}) provides much more robust estimates of $\rho$ than the usual Pearson estimator. The following result from \cite{rupp} can be useful for fitting the Gaussian and $t$ Copula.
Let $Y=(Y_{1},...,Y_{n})^{T}$ have a meta-Gaussian distribution\footnote{A meta-Gaussian distribution is any distribution that has the Gaussian copula} with continuous marginal distribution and copula $C_{P}^{\text{Gau}}$ and let $P_{i,j}=\left[P\right]_{i,j}$: Then,
\begin{equation}
\rho_{\tau}\left(Y_{i},Y_{j}\right)= \frac{2}{\pi} \text{arcsin}(P_{i,j})
\label{ken2}
\end{equation}
and 
\begin{equation}
\rho_{s}(Y_{i},Y_{j})=\frac{6}{\pi} \text{arcsin} \left( \frac{P_{i,j}}{2}\right)  \approx P_{i,j}
\label{spear2}
\end{equation}
If instead $Y$ has a meta-$t$ distribution\footnote{A meta-$t$ distribution is any distribution that has the $t$-copula} with continuous univariate marginal distributions and copula $C_{\nu,P}^{t}$ then only (\ref{ken2}). For more details about Kendall's tau and Spearman's rho and their estimation, we refer to \cite{martinH}, \cite{kendall}, \cite{kruskal}, \cite{lehmann}. 
%

%
\section{Synthetic data using Copulas} 
\subsection{Generating Synthetic data with numerical features}
\label{confea}
Our aim in this section is to propose a copula-based methodology to generate data containing numerical (continuous) features. We will first isolate the dependency structure of the dataset by copulas. As defined above, we will focus on the elliptic copulas and in this context the dependency between the features can be explained by the linear correlation matrix. Second, we will estimate the marginal distributions of the different features and then we can use the inverse transform method to genate them.  Recall the Inverse Transform method.


\medskip
\newpage

\textbf{Inverse Transform Method}

Let $F(x)$, $x \in \mathbf{R}$ denote any cumulative distribution function (cdf)(continuous or discrete). Let $F^{-1}(y)$, $y \in [0,1]$ denote the inverse function defined by 
\begin{equation}
F^{-1}(y)=\text{min}\left\lbrace x: F(x)\geq y \right\rbrace, y \in [0,1]
\end{equation}
Define $X=F^{-1}(U)$ where $U$ has the continuous uniform distribution over the interval $(0,1)$. Then $X$  is distributed as $F$, that is, $P(X\leq x)=F(x)$, $x \in \mathbf{R}$.

Now, let $X=(X_{1},...,X_{n})$ be the real dataset and one would like to produce substitute data closely resembling to the real data in terms of marginal and joint distributions. Let $P=cor(X)$ be its corresponding Pearson correlation matrix. A robust way to estimate the Pearson correlation $P_{i,j}=\left[P\right]_{i,j}$ is to estimate the Kendall's tau in the first step and to use (\ref{ken2}). Using the Inverse Transform Method and the Sklar's theorem, we can generate the synthetic dataset by the following steps:
\begin{enumerate}
\item Generate the elliptic copulas $C$ such as the Gaussian or t copulas using the correlation matrix $P$, as explained in section \ref{pop}. Let $U=(U_{1},...,U_{n})$ be the generated set of Uniform random variables that has the joint cdf, the copula $C$.
\item Find the marginal cumulative distributions, $G_{i}$ for each feature. In the case $G_{i}$ is not known, one can estimate them by the empirical cumulative distribution function (ecdf), $\hat{G_{i}}$
\item Generate multivariate $\eta=(\eta_{1},...,\eta_{n})$ such that $\eta_{i}=\hat{G_{i}}^{-1}(U_{i})$. $\eta$ is in fact the synthetic data that shares the same distribution (marginal or joint) with the real dataset $X$. $\eta$ has a Meta-elliptical distribution since it is generated by the elliptical copulas.
\end{enumerate}
\subsection{Generating Synthetic data with categorical features}
\label{catfea}
Most of the research on the copulas has been based upon the assumption that the univariate margins are continuous. Without this assumption, many results are no longer valid and the interaction between the copula and the margins is much less well understood. In fact, when discontinuities are allowed, there exists a whole class of copulas which comply with the famous Sklar's theorem, see \cite{schw}. This is generally not a problem in applied contexts, as researchers use copulas because the joint distribution is either not known or is difficult to work with. A much more serious problem with discontinuous margins is that estimates of the dependence parameters are biased and therefore our generated dataset is far to keep the dependence structure of the original dataset. In fact, when the margins are discontinuous, they present jumps that cause their inverse to have plateaus and this potentially leads to biased estimates, see \cite{genest}.

In this context, to generate categorical features using copulas, we propose to encode the categorical features into continuous features first and then we can apply the method developed in (\ref{confea}). The process is the following:
\begin{enumerate}
\item Consider the dataset with $n$ rows and $m$ columns; we suppose that the $m$ columns are exclusively categorical features denoted by $C_{k}$ with $1\leq k\leq m$. Let $C_{k}^{j}$ the $j$ value of the categorical feature $C_{k}$ with $1\leq j\leq n$. In other words, $n$ is the size of the dataset. 
\item Let $m_{k}$ the number of levels (groups) in the feature $C_{k}$. We estimate the proportions for each level $i$ in the categorical feature $C_{k}$. Let $\hat{P}_{k}^{i}$ the estimated proportion; we have $\sum_{i}^{m_{k}} \hat{P}_{k}^{i}=1$.
 \item The confidence interval for the proportion $\hat{P}_{k}^{i}$ is computed with the following formula
 $$ \text{CI}_{k}^{i}=\hat{P}_{k}^{i} \pm z\sqrt{\frac{\hat{P}_{k}^{i}(1-\hat{P}_{k}^{i})}{n}}$$
 where $z$ is the $Z$-value for the desired confidence level (e.g. $z=1.96$ for $95\%$ confidence) and $n$ is assumed sufficiently large.
 \item To convert each level $i$ to  continuous value, we replace the estimated proportion, $\hat{P}_{k}^{i}$, with a value $\hat{V}$ sampled from a normal distribution with mean $\hat{P}_{k}^{i}$ and variance $\frac{\hat{P}_{k}^{i}(1-\hat{P}_{k}^{i})}{n}$. So, for each feature $C_{k}$ one obtains $n$ numerical values $\hat{V}_{k}^{j}$.
\item Getting the numerical features, one can generate the full dataset by applying the method based on the copulas that is described in section (\ref{confea}). 
We denote $\hat{G}_{k}^{j}$ the simulated value of $\hat{V}_{k}^{j}$ by taking account the dependency between the numerical features.
\item After transforming the categorical variables into numerical variables and generating them, the main question is how to convert the generated numerical features into categorical features in order to find the structure of the original dataset? 

We propose the following steps:
\item For each $\hat{G}_{k}^{j}$, we calculate 
$$D_{k}^{i,j}=\lvert\hat{G}_{k}^{j}-\hat{P}_{k}^{i}\rvert\,\, \text{for all} \,\,\hat{P}_{k}^{i} \,\, \text{with} \,\,  1\leq i \leq m_{k}$$ 
\item Let $a_{k}^{j}=\text{min}_{i}(D_{k}^{i,j})$; find the level $i$ or set of levels $S_{k}$ for which $D_{k}^{i,j}=a_{k}^{j}$. In other words, we are looking for the levels $i$ for which the proportions are close to the simulated value $\hat{G}_{k}^{j}$.
\item If two or more levels are found and $C_{k}^{j} \in S_{k}$ then we select $C_{k}^{j}$. If $C_{k}^{j} \notin S_{k}$, then we select randomly one level in $S_{k}^{i}$.
\end{enumerate}

\begin{table}[t]
\caption{Table with two categorical features: vehicle type and vehicle color}
\label{table:1}
\vskip 0.15in
\begin{center}
\begin{small}
\begin{sc}
 \begin{tabular}{c c c} 
\toprule
& VEHICLE & COLOR\\
\midrule
 1 & CAR & BLUE \\ 
 2 & BUS & GREEN \\
 3 & BICYCLE & GREEN \\
 4 & BUS & BLUE  \\
 5 & CAR & GREEN \\  
 6 & BUS & BLUE \\  
 7 & CAR & GREEN \\  
  8 & BICYCLE & BLUE \\  
  9 & BICYCLE & BLUE \\  
  10 & BUS & GREEN \\ [1ex] 
\bottomrule
 \end{tabular}%
\end{sc}
\end{small}
\end{center}
\vskip -0.1in
\end{table}

Let illustrate the above algorithm with the following example in Table \ref{table:1}. In this table, we have 2 categorical features of which the first feature relates to the vehicule type and the second feature relates to the color of the vehicle. 

Our goal is to give an idea of how to simulate this table using the above method. Of course, we have only 10 observations and it would be difficult to get a good estimate and therefore a good simulation. Here, $m=2$ and $n=10$. The numbers of levels in the feature $C_{1}$ and in the feature $C_{2}$ are $m_{1}=3$ and $m_{2}=2$ respectively. In the feature $C_{1}$, we denote $1=\text{CAR}$, $2=\text{BUS}$, $3=\text{BICYCLE}$. In the feature $C_{2}$, $1=\text{BLUE}$, $2=\text{GREEN}$. We have the following estimations of the proportions: 
\begin{equation*}
\begin{aligned}
P_{1}^{1}=3/10, \,\,P_{1}^{2}=&4/10, \,\, P_{1}^{3}=3/10, \,\, \\
&P_{2}^{1}=5/10,\,\,P_{2}^{2}=5/10
\end{aligned}
\end{equation*}


Based on the estimated proportions, one can generate numerical features by applying the steps $4$ to $5$ of the above algorithm. Suppose that we obtain the following numerical table

Now, the last step is to convert this table into categorical one. Taking two values in the table, say $\hat{G}_{1}^{1}=7/10$ and $\hat{G}_{2}^{3}=1/3$, let's apply the conversion process.
\begin{itemize}
\item For $\hat{G}_{1}^{1}=7/10$

 $ P_{1}^{1}=3/10, \,\,P_{1}^{2}=4/10, \,\, P_{1}^{3}=3/10$ and $D_{1}^{1,1}=4/10$, $D_{1}^{2,1}=3/10$, $D_{1}^{3,1}=4/10$; we have  $a_{1}^{1}=D_{1}^{2,1}=3/10$ and therefore $\hat{G}_{1}^{1}$ will be replaced by the level $2$ of the feature $C_{1}$ which is \textbf{BUS}.
\item For $\hat{G}_{2}^{3}=1/3$

 $P_{2}^{1}=5/10,\,\,P_{2}^{2}=5/10$ and $D_{2}^{1,3}=1/6$, $D_{2}^{2,3}=1/6$; we have $a_{2}^{3}=1/6$ and $S_{2}=\left\lbrace \text{BLUE},\text{GREEN} \right\rbrace$, As $C_{2}^{3}=\text{GREEN} \in S_{2}$, then $\hat{G}_{2}^{3}$ will be replaced by \textbf{GREEN}.
\end{itemize}

\begin{table}[t]
\vskip 0.15in
\begin{center}
\begin{small}
\begin{sc}
 \begin{tabular}{c c c} 
\toprule
 & $\hat{G}_{1}$ & $\hat{G}_{2}$\\ 
\midrule
1 & 7/10 & 3/4 \\ 
 2 & 4/7 & 7/8 \\
 3 & 1/100 & 1/3 \\
 4 & 1/80 & 2/9  \\
 5 & 1/2 & 3/5 \\ 
 6 & 1/6 & 2/3 \\ 
 7 & 1/6 & 3/11 \\ 
  8 & 3/5 & 1/101 \\
  9 & 7/10 & 5/9 \\
  10 & 3/11 & 1/3 \\ 
\bottomrule
 \end{tabular}%
\end{sc}
\end{small}
\end{center}
\vskip -0.1in
\end{table}
Applying the same process to the rest of the table gives the final categorical Table \ref{table:2}.
\section{Application to Datasets} 
To access the ability of our model to generate synthetic data, we focused on a public and private datasets. For publicly available datasets, we used a Kaggle dataset; this datasets contains transactions made by credit cards in September $2013$ by European cardholders and presents transactions that occurred in two days. It contains $30$ numerical features and among them $28$ are the results of PCA transformations. The only features that have not been transformed by the PCA are the "Time" and the "amounts". Feature "Time" contains the seconds elapsed between each transaction and the first transaction in the datasets. This feature being deterministic, we remove it from the original dataset because it can bias our simulation by copulas. The second dataset is private and consists of transactions for customers of a financial institution. The private datasets include $15$ features, two of which are categorical. Our goal is to generate both datasets by the proposed methodology. 

\begin{table}[t]
\caption{Generating Synthetic table based on table \ref{table:1}}
\label{table:2}
\vskip 0.15in
\begin{center}
\begin{small}
\begin{sc}
 \begin{tabular}{c c c} 
\toprule
& VEHICLE & COLOUR\\
\midrule
 1 & BUS & BLUE \\ 
 2 & BUS & GREEN \\
 3 & BICYCLE & GREEN \\
 4 & CAR & BLUE  \\
 5 & BUS & GREEN \\  
 6 & CAR & BLUE \\ 
 7 & CAR & GREEN \\  
  8 & BUS & BLUE \\  
  9 & BUS & BLUE \\  
  10 & CAR & GREEN \\ [1ex] 
\bottomrule
 \end{tabular}%
\end{sc}
\end{small}
\end{center}
\vskip -0.1in
\end{table}

\subsection{Generating the Kaggle datasets} 
The method described in section (\ref{confea}) is applied to Kaggle datasets because these datasets include only numerical features. For simplicity, we focus on Gaussian copulas. As noted above, we removed the feature "Time" from the original dataset because this feature is not random. Our results are compared to those generated by autoencoders and SMOTE. 


\medskip

\textbf{Autoencoders}

An autoencoder (AE) is an artificial neural network (or neural network) used to learn a representation of a set of data, \cite{liou}. It has an input layer, an output layer and one or more hidden layers with the purpose of reconstructing its original input. There are two parts of an autoencoder; the encoder and the decoder. The encoder is the subnetwork from the input layer to the middle layer and reduces the dimension of the input data to learn simplified representation of it. The decoder is the subnetwork from the middle layer to the output layer and reconstructs the input data, \cite{jin1}.

To generate synthetic datasets, we consider an autoencoder with three hidden layers of $16$, $8$ and $16$ hidden neurons respectively. For the parameters, we fix a batch size of $128$, $100$ epochs and learning rate of 1e-7. 

\medskip

\textbf{Synthetic Minority Over Sampling (SMOTE)}

Synthetic Minority Over Sampling (SMOTE) synthesizes new data instances between existing (real) instances and was developed to address the problems in applying classification algorithms to unbalanced datasets. Proposed in $2002$ by \cite{chaw} it is now an established method, with over $85$ extensions of the basic method reported in specialised literature. A way to visualise how the basic concept works is to imagine drawing a line between two existing instances. SMOTE then creates new synthetic instances somewhere on these lines. 

Considering a sample $x_{i}$, a new sample will be generated considering its $k$ nearest neighbours. The steps are the following:
\begin{itemize}
\item Take the difference between the feature vector sample under consideration and one of those $k$ nearest neighbours, $x_{ki}$.
\item Multiple the difference by a random number $\gamma$ between $0$ and $1$ and add this to the sample vector to generate a new sample $x_{\text{new}}$
$$ x_{\text{new}}=x_{i} + \gamma (x_{ki}-x_{i}) $$
\end{itemize}
Variations of the technique have been developed by varying the way the samples  and their neighbours are selected before generating new ones. For more details, see \cite{bai}, etc... 

To access the quality of the algorithms for generating synthetic datasets, we used two criteria of comparison: 
\begin{enumerate}
\item Determine whether the relationships between the variables in the original data are preserved in the synthetic data. 
\item Determine whether the marginal distributions of the features in the original dataset are preserved in the synthetic dataset. 
\end{enumerate}

\subsubsection{Testing the quality of dependency or association on generated synthetic datasets} 
To evaluate the quality of the synthetic datasets to preserve the dependency structure of the Kaggle datasets we proceed as follows:

\begin{itemize}
\item We evaluate the pairwise correlation between the $n$ variables of the real datasets. Let 
$P_{\text{real}}$ be the correlation matrix
\item We evaluate the pairwise correlation between the $n$ variables of the synthetic datasets. Let $P_{\text{syn}}$ be the correlation matrix
\item We calculate $P_{\text{diff}}=P_{\text{real}}-P_{\text{syn}}$
\item We compute $\mu_{\text{diff}}$, that is the mean of the absolute values of all elements of $P_{\text{diff}}$. $\mu_{\text{diff}}$ captures the proximity in terms of correlation between real data and synthetic data. Therefore lower values of $\mu_{\text{diff}}$ represents higher accuracy. $\mu_{\text{diff}}$ serves as a concise quantitative measure of synthetic data quality
\end{itemize}

\begin{figure}
\begin{multicols}{2}
\includegraphics[width=\linewidth]{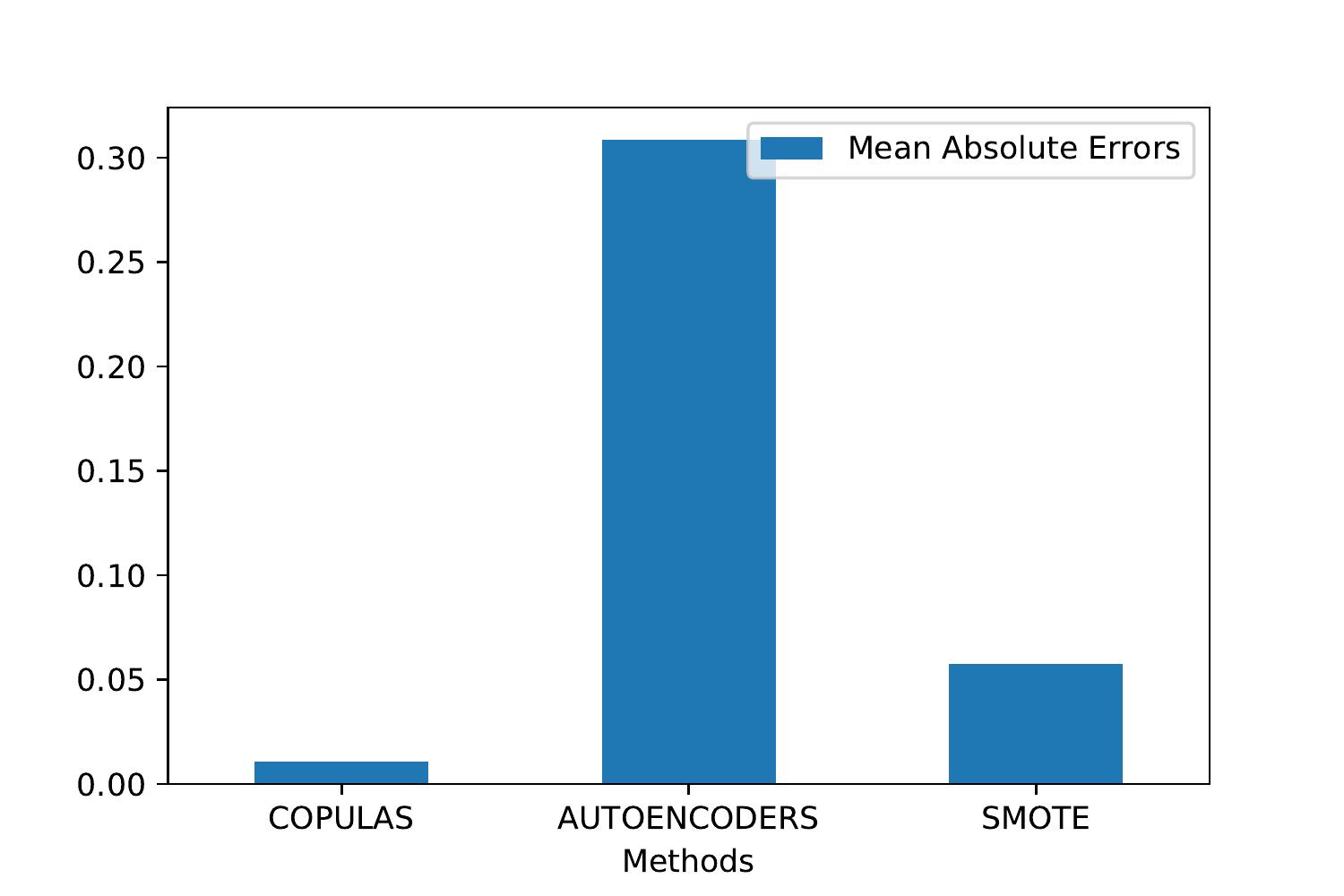}\par 
\includegraphics[width=\linewidth]{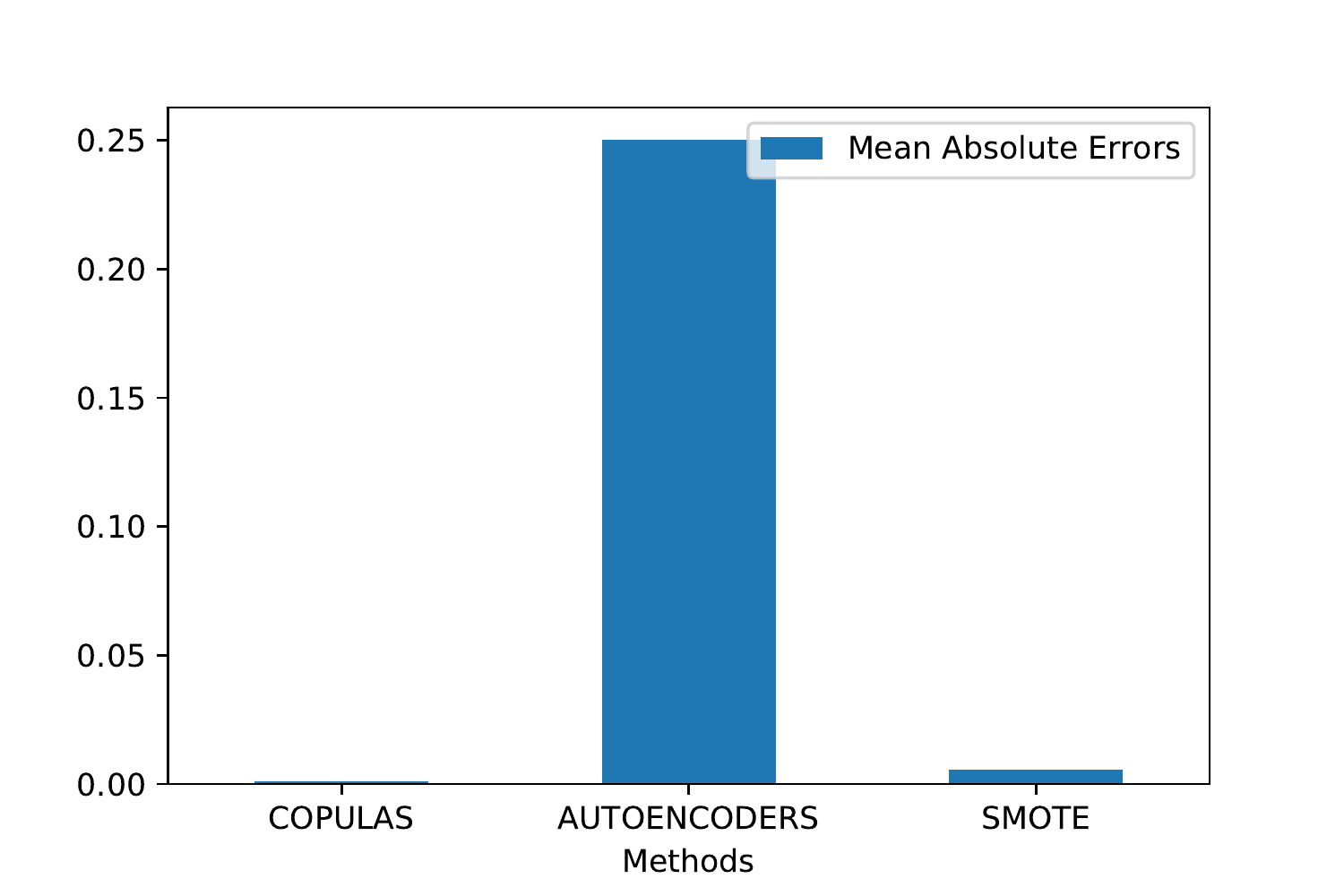}\par 
\includegraphics[width=\linewidth]{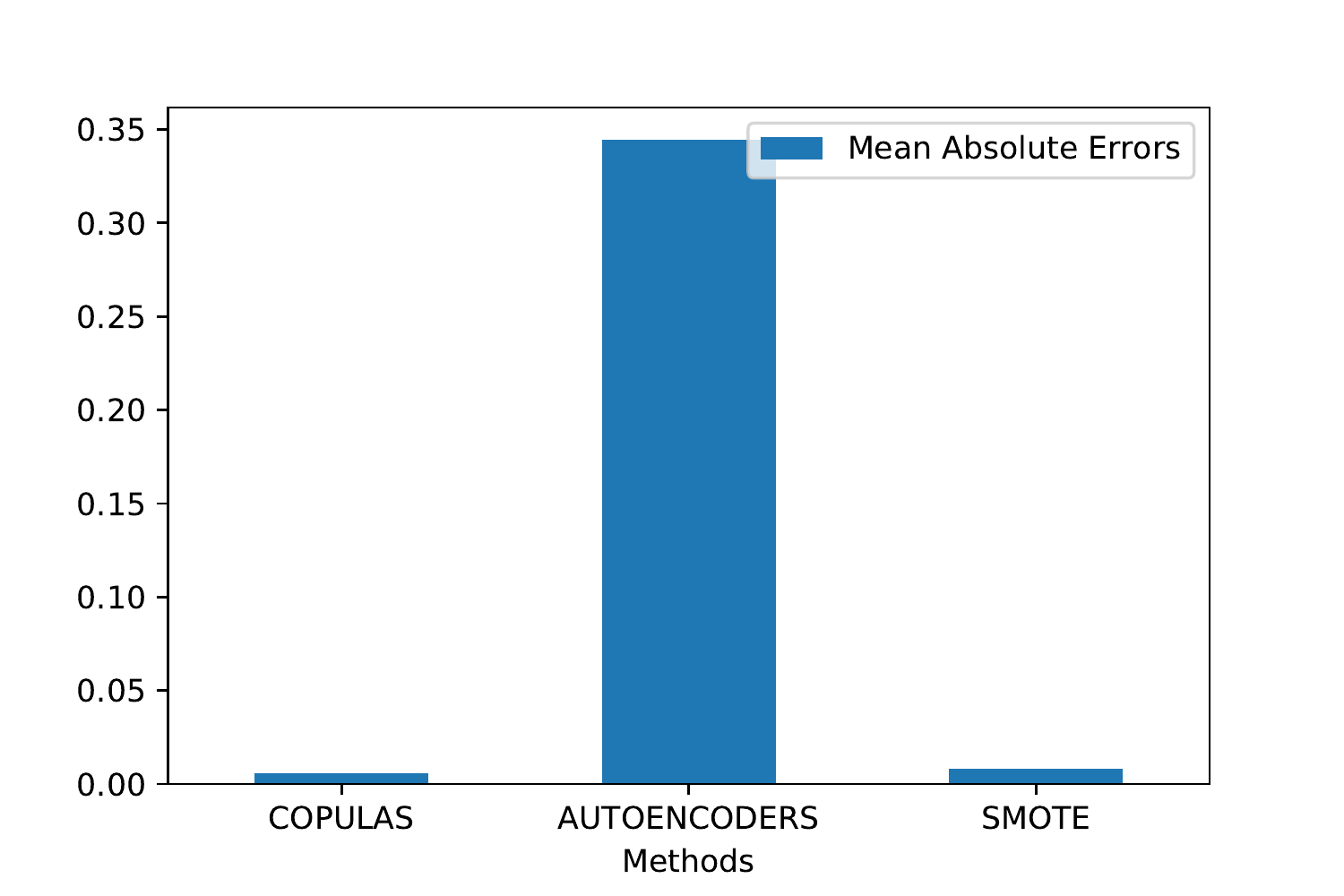}\par
\end{multicols}
\caption{Mean absolute errors on the correlation of the datasets generated by the copulas, the autoencoders and the SMOTE methods. We consider in (a) the Pearson correlation, in (b) the Kendall's tau and in (c) the Spearman correlation. }
\label{fig7}
\end{figure}

We evaluate the quality in terms of dependency of the synthetic datasets generated by the copulas, the autoencoders and the SMOTE. We consider the Pearson correlation, the rank correlation such as the Kendall's tau and the Spearman correlation. The results are shown in Figure \ref{fig7}. We notice that the copulas method produced high-quality synthetic samples because its $\mu_{\text{diff}}$ shows lower values for the $3$ types of correlation. This means that the dependency structure of the real datasets (Kaggle datasets) was closely mirrored by the dependency structure of the copulas synthetic datasets. The copulas method was followed by the SMOTE method which presents very close results for the Spearman correlation as $\mu_{\text{diff}}=0.005$ for the copulas and $\mu_{\text{diff}}=0.008$ for the SMOTE. On the other hand, the autoencoders method produced the least accurate result because they present higher values of $\mu_{\text{diff}}$. It is important to point out that for the implementation of the autoencoders, we use the common default parameters. Their performances may be improved with further hyperparameter optimisation.

\subsubsection{Testing the quality of marginal distribution on generated synthetic datasets} 

\begin{figure}
\includegraphics[width=\linewidth]{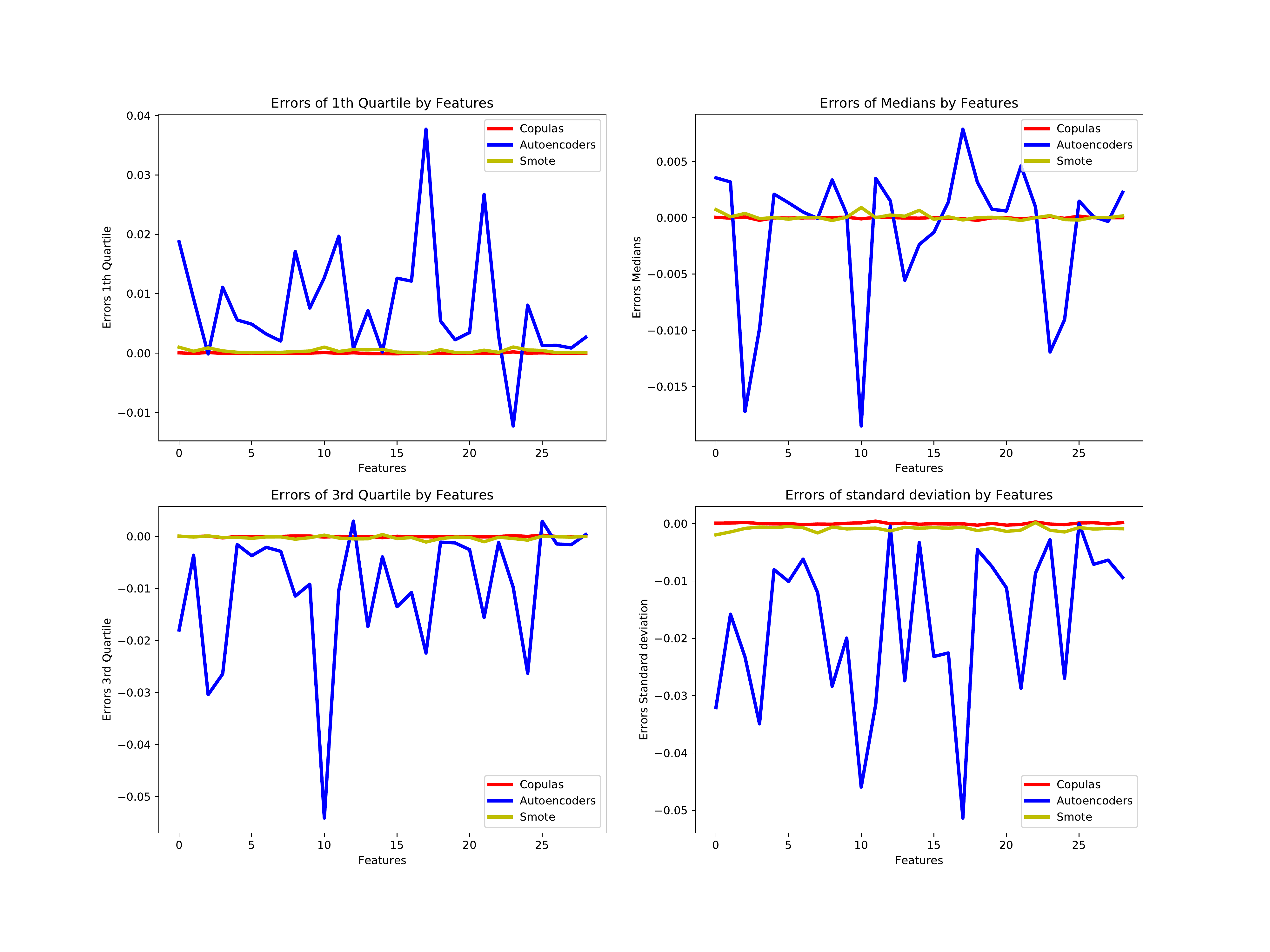}
\caption{Errors on the descriptive statistics on the feature variables between the $3$ synthetic and Kaggle datasets. }
\label{fig8}
\end{figure}

\begin{figure}
\includegraphics[width=\linewidth]{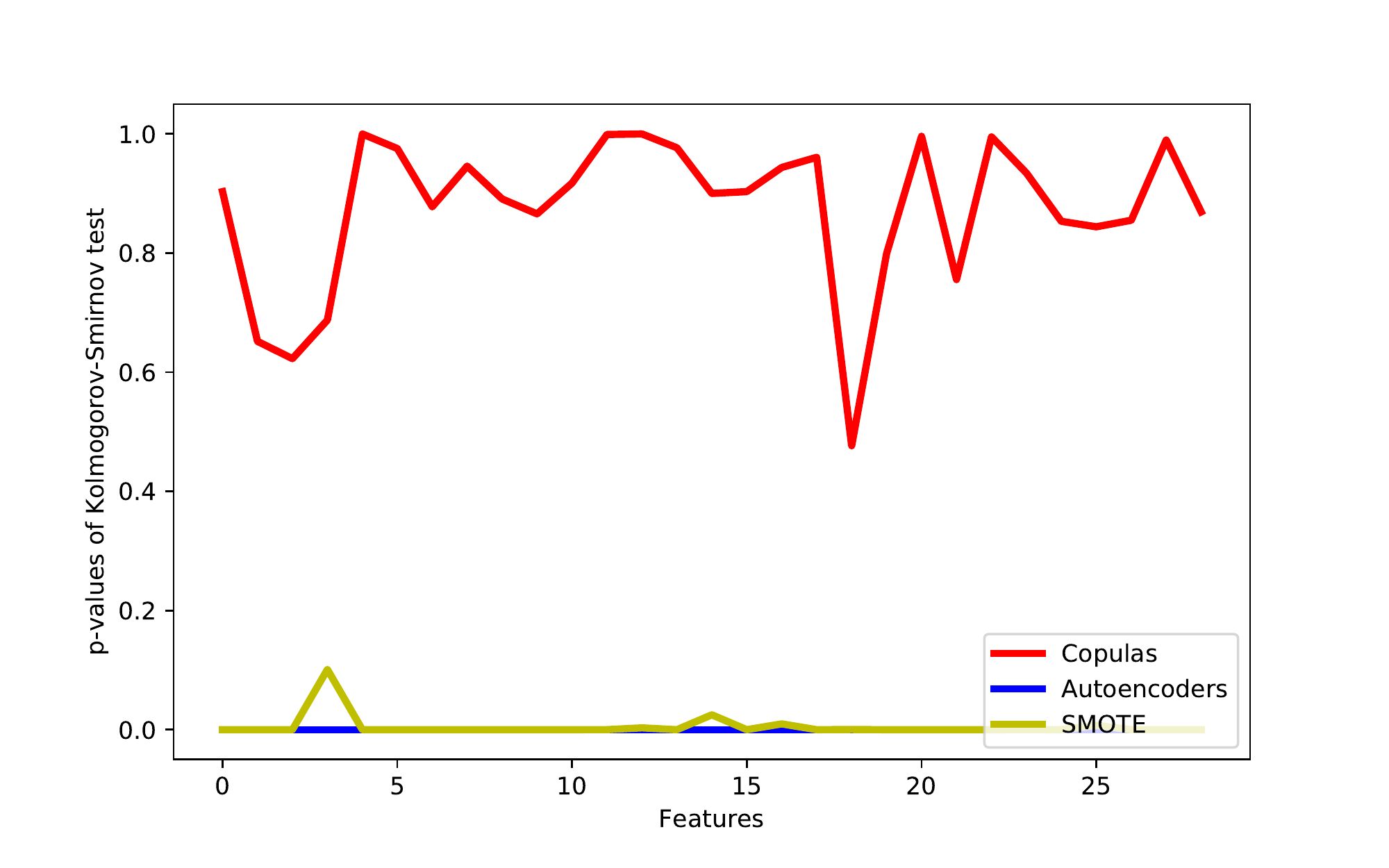}
\caption{p-values of the Kolmogorov-Smirnov (K-S) tests on the feature variables between the Kaggle and the generated datasets }
\label{fig9}
\end{figure}

To evaluate how the generated datasets preserve the marginal distributions of the Kaggle datasets, we estimate some descriptive statistics such as the quartiles and the standard deviation on the feature variables of the Kaggle and synthetic datasets. Next, we calculate the errors between these statistics that are shown in Figure \ref{fig8}. We also proceed to the Kolmogorov-Smirnov (K-S) test that compares the empirical cumulative distribution functions between the  feature variables of the Kaggle and synthetic datasets. Figure \ref{fig9} shows the results of the p-values. Lower values of p-values ($< 5\%$) means that we reject the hypothesis that the distribution is the same. The analysis of the two figures will capture the proximity in terms of marginal distribution between the kaggle and the synthetic datasets.

The analysis of Figure \ref{fig8} reveals significant errors between the statistics of the autoencoders methods and the Kaggle datasets while the errors are in general close to zeros for the copulas and the SMOTE. This means that both copulas and the SMOTE are able to preserve the statistic measures of the Kaggle datasets during the generation of datasets. Analysis of the p-values of the K-S test in Figure \ref{fig9} shows that the hypothesis of same distribution between the variables of the Kaggle and the copulas datasets can not be rejected at the significance level of $5\%$. The small values for the p-values of the autoencoders and the SMOTE generally leads to reject the assumption of same distribution on the feature  variables between these datasets and the Kaggle dataset. We can conclude that among the $3$ methods, the copula method is the most efficient in generating synthetic datasets while preserving correlations and marginal distributions.


\subsection{Generating the private categorical datasets} 

Private datasets are provided by Netguardians and consist of transactions for customers of a financial institution and includes 15 features of which 2 are categorical. It contains a total of $1,000$ transactions made in various currencies to different destinations. For confidentiality reasons the names of the financial institution are not mentioned. Our goal is to generate these datasets using the copulas method, autoencoders and SMOTE and compare their performance. To assess performance, we will especially focus on how the generated datasets capture the association between the two categorical features. 

The datasets generation process  here consists of performing some transformations to the categorical features to obtain numerical features before applying the $3$ datasets generation methods. Then, we apply the second phase of the transformation methods to obtain the categorical features as described in section (\ref{catfea}). The $2$ categorical features concern the currency and the destination on the transactions for certain clients of the financial institution.  Table \ref{table:3} shows the cross-tabulation of two categorical variables in the datasets. The " Destination" variable has $8$ levels and the variable " currency" has 6 levels and the columns define the number of transactions that are made in different currencies for a given destination. We transform the two categorical features into numerical ones, then we apply the $3$ synthetic generation methods to these datasets following the algorithms in section (\ref{catfea}). The results of the cross-tabulations between the $2$ categorical generated features are shown in Tables \ref{table:4} to  \ref{table:6}.
%

\begin{figure}[H]
\includegraphics[width=\linewidth]{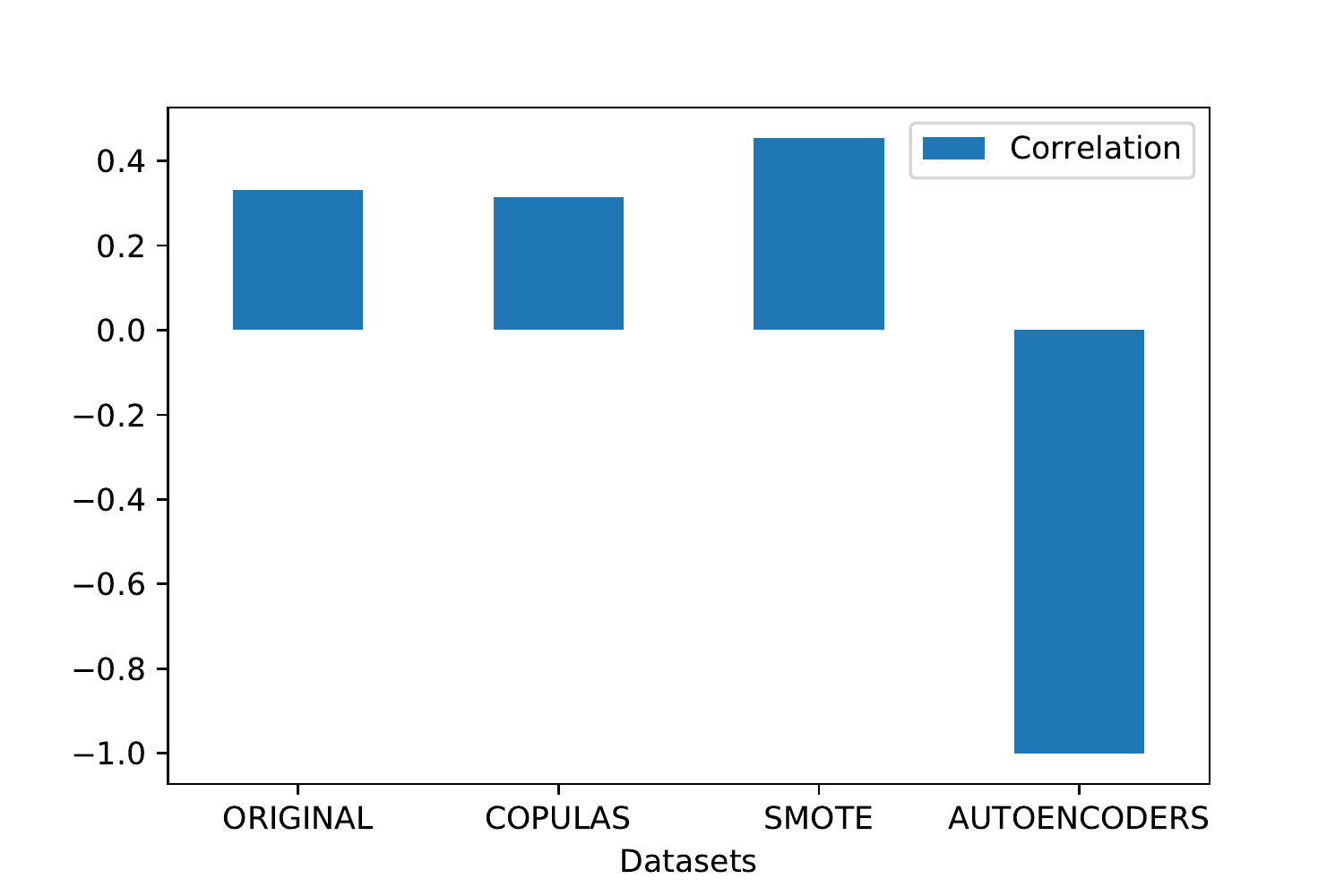}
\caption{The Kendall's tau between the two transformed continuous features in the datasets}
\label{fig10}
\end{figure}

Table \ref{table:4} shows that the copulas method focused on the levels of the two categorical features that are more associated and neglected the levels that are less associated. In other words, the copula method during the process of generating synthetic datasets reproduces essentially the characteristics of the levels of the categorical features that are more associated. In Table \ref{table:5}, it seems that the SMOTE method tends to keep most of the levels for the two features but there is the presence of noises in the distribution which will probably reduce the association between the two features. In Table \ref{table:6}, the autoencoders method focuses only on one level of the currency and its repartition through the destinations changes significantly compared to the original datasets. The Chi-squared test for independence is applied to the contingency tables and shows that the independence between the categorical features is rejected for all datasets except the one generated by autoencoders. Finally, Figure \ref{fig10} shows the Kendall's tau correlation between the continuous features in the synthetic datasets after the transformation of the categorical ones. It reveals that among the $3$ methods, the copulas method captures more closely the correlation in the original datasets between the two continuous features. The autoencoders method shows a high negative correlation that differs significantly from the original datasets. 

\section{Conclusion}
We present a method of generating synthetic data based on copulas and in this context it is shown how to deal with categorical variables. Our model is experimented on Kaggle and private datasets and its performance is compared to that of the SMOTE and autoencoders models. The results show that the copulas approach clearly outperforms other approaches because it better preserves the dependency structure and marginal distributions of the original datasets. The autoencoders method produces the least accurate results but its performance could eventually be improved with further hyperparameters optimization. It is important to note that this paper describes the first steps in our data synthesis work. For the categorical variables generation, we conducted our experiments on a private dataset consisting of approximately $1,000$ records and $15$ variables, of which we focused on $2$ categorical variables. We are aware that in many cases it is necessary to synthesise categorical datasets with hundreds of thousands records and several hundreds of categorical variables. We therefore acknowledge that further experimentation is required to draw definitive conclusions with respect to the ability of the tested algorithms to generate synthetic data especially in the context of categorical variables. This is left for future work.

\begin{table*}
\caption{Cross-tabulation between the features Currency and Destination in the original datasets}
\label{table:3}
\vskip 0.15in
\begin{center}
\begin{small}
\begin{sc}
 \begin{tabular}{c c c c c c c c c} 
\toprule
 {Currency}{Destination} & Dst1 & Dst2 & Dst3 & Dst4 & Dst5 & Dst6 & Dst7 & Dst8\\
\midrule
 Cur1 & 0 & 3 & 0 & 0 & 0 & 0 & 0 & 0 \\ 
 Cur2 & 550 & 0 & 5 & 0 & 45 & 0 & 0 & 0 \\
 Cur3 & 100 & 0 & 1 & 1 & 0 & 0 & 198 & 0 \\
 Cur4 & 2 & 0 & 0 & 0 & 0 & 91 & 0 & 1  \\
 Cur5 & 0 & 0 & 0 & 0 & 0 & 0 & 0 & 1 \\  
 Cur6 &  0 & 2 & 0 & 0 & 0 & 0 & 0 & 0 \\ 
\bottomrule
 \end{tabular}%
\end{sc}
\end{small}
\end{center}
\vskip -0.1in
\end{table*}
%
%
\begin{table*}
\caption{Cross-tabulation between the features Currency and Destination in the syntehtic datasets generated by the copulas}
\label{table:4}
\vskip 0.15in
\begin{center}
\begin{small}
\begin{sc}
 \begin{tabular}{c c c c c c c c} 
\toprule
{Currency} {Destination} & Dst1 & Dst2 & Dst3 & Dst4 & Dst5 & Dst6 & Dst7\\
 \midrule
 Cur2 & 452 & 1 & 1 & 0 & 17 & 7 & 21 \\ 
 Cur3 & 120 & 0 & 0 & 0 & 0 & 0 & 257 \\
 Cur4 & 1 & 0 & 0 & 1 & 1 & 120 & 1 \\
\bottomrule
 \end{tabular}%
\end{sc}
\end{small}
\end{center}
\vskip -0.1in
\end{table*}
\begin{table*}
\caption{Cross-tabulation between the features Currency and Destination in the syntehtic datasets generated by SMOTE}
\label{table:5}
\vskip 0.15in
\begin{center}
\begin{small}
\begin{sc}
 \begin{tabular}{c c c c c c c c} 
\toprule
 {Currency}{Destination} & Dst1 & Dst2 & Dst3 & Dst4 & Dst5 & Dst6 & Dst7\\ 
\midrule
 Cur1 & 1 & 0 & 0 & 0 & 0 & 0 & 2 \\ 
 Cur2 & 450 & 1 & 0 & 0 & 12 & 33 & 102 \\
 Cur3 & 153 & 3 & 3 & 0 & 20 & 44 & 82 \\
 Cur4 & 24 & 0 & 0 & 2 & 6 & 19 & 30  \\
 Cur5 & 0 & 0 & 0 & 0 & 0 & 1 & 0 \\  
 Cur6 &  0 & 0 & 1 & 0 & 0 & 1 & 1\\ 
\bottomrule
 \end{tabular}%
\end{sc}
\end{small}
\end{center}
\vskip -0.1in
\end{table*}
\begin{table*}
\caption{Cross-tabulation between the features Currency and Destination in the syntehtic datasets generated by the autoencoders method}
\label{table:6}
\vskip 0.15in
\begin{center}
\begin{small}
\begin{sc}
 \begin{tabular}{c c c c c c c c} 
\toprule
{Currency}{Destination} & Dst2 & Dst3 & Dst4 & Dst5 & Dst6 & Dst7 & Dst8\\
\midrule
 Cur2 & 57 & 127 & 559 & 8 & 24 & 187 & 38 \\ 
\bottomrule
 \end{tabular}%
\end{sc}
\end{small}
\end{center}
\vskip -0.1in
\end{table*}


\clearpage
\cleardoublepage

\bibliographystyle{icml2021}
\bibliography{toto}

\end{document}